\documentclass[10pt,twocolumn,letterpaper]{article}

\usepackage{cvpr}              %

\usepackage[dvipsnames]{xcolor}
\newcommand{\red}[1]{{\color{red}#1}}

\newcommand{\TODO}[1]{\textbf{\color{red}[TODO: #1]}}
\renewcommand{\TODO}[1]{}

\usepackage{amsmath}
\usepackage{indentfirst}
\usepackage{booktabs}
\usepackage{multirow}
\usepackage{caption}
\usepackage{subcaption}
\usepackage{multirow}
\usepackage{gensymb}
\usepackage{nicefrac}
\usepackage{algorithm}
\usepackage{stmaryrd}
\usepackage[noend]{algpseudocode}
\usepackage{algorithmicx}
\newcommand{\norm}[1]{\left\lVert#1\right\rVert}
\usepackage{multirow,makecell}
\newcommand{\cross}{\times}
\DeclareMathOperator*{\argmax}{arg\,max}
\DeclareMathOperator*{\argmin}{arg\,min}

\usepackage{tikz}
\usepackage{mathtools}
\newcommand{\myarrow}[1][-45]{%
  \mathrel{%
    \text{$
     \begin{tikzpicture}[baseline = -0.5ex]
       \node[inner sep=0pt,outer sep=0pt,rotate = #1] (a) at (0,0)  {$\xrightarrow{}$};
    \end{tikzpicture}
    $}%
  }%
}%

\definecolor{cvprblue}{rgb}{0.21,0.49,0.74}
\usepackage[pagebackref,breaklinks,colorlinks,citecolor=cvprblue]{hyperref}

\def\paperID{480} %
\def\confName{CVPR}
\def\confYear{2024}

\title{Multi-Session SLAM with Differentiable Wide-Baseline Pose Optimization}

\author{Lahav Lipson\\
Princeton University\\
{\tt\small llipson@princeton.edu}
\and
Jia Deng\\
Princeton University\\
{\tt\small jiadeng@princeton.edu}
}

\begin{document}
\newcommand*{\Simt}{\text{Sim}(3)}

\maketitle

\begin{abstract}
We introduce a new system for Multi-Session SLAM, which tracks camera motion across multiple disjoint videos under a single global reference. Our approach couples the prediction of optical flow with solver layers to estimate camera pose. The backbone is trained end-to-end using a novel differentiable solver for wide-baseline two-view pose. The full system can connect disjoint sequences, perform visual odometry, and global optimization. Compared to existing approaches, our design is accurate and robust to catastrophic failures. Code is available at \url{https://github.com/princeton-vl/MultiSlam_DiffPose}\vspace{-5mm}%
\end{abstract}

\section{Introduction}

Simultaneous Localization and Mapping (SLAM) is the task of estimating camera motion and a 3D map from video. The standard setup assumes a single continuous video. However,  video data in the wild often consists of not a single continuous stream, but rather multiple disjoint sessions, either deliberately such as in collaborative mapping when multiple robots perform joint rapid 3D reconstruction, or inadvertently due to visual discontinuities in the video stream which can result from camera failures, extreme parallax, rapid turns, auto-exposure lag, dark areas, or extreme occlusion by dynamic objects. Handling such disjoint videos is important for many applications in AR and robotics, and gives rise to the task of Multi-Session SLAM.\looseness=-1

In Multi-Session SLAM, the input consists of multiple disjoint video sequences and the goal is to estimate camera poses for all video frames under a single global reference. This is in contrast to ``single-video SLAM'', whose input is a single continuous video. In this work, we focus on the monocular, RGB-only Multi-Session SLAM setting. 

Several approaches have been proposed to deal with Multi-Session SLAM, however existing solutions typically require additional sensor data in order to remove gauge freedoms and make tracking easier~\cite{qin2018vins, labbe2019rtab, schneider2018maplab}. Only a small number of methods, notably CCM-SLAM~\cite{schmuck2019ccm} and ORB-SLAM3~\cite{orbslam3}, support Multi-Session SLAM from monocular video alone, due to the difficulty of aligning disjoint sequences under the 7-DOF gauge freedoms in monocular video. However, these approaches are based on classical feature descriptors, making them less accurate on average compared to recent designs based on deep networks. 

In the standard SLAM setting, Teed and Deng~\cite{teed2021droid} proposed to use a deep optical flow network (RAFT~\cite{teed2020raft}) to track 2D motion, while jointly updating camera poses with a bundle adjustment layer. The method, DROID-SLAM, is accurate and avoids tracking failures, but the design assumes a continuous video stream and is not capable of the wide-baseline matching and non-local optimization necessary for Multi-Session SLAM. Deep Patch Visual Odometry~\cite{dpvo} (DPVO) introduced a sparse visual-odometry-only analog of DROID-SLAM which achieves similar accuracy on single-video VO, but at much lower cost. However, DPVO also does not support Multi-Session SLAM for the same reasons as DROID-SLAM.\looseness=-1 %

\begin{figure}[t]
    \centering
    \includegraphics[width=\columnwidth]{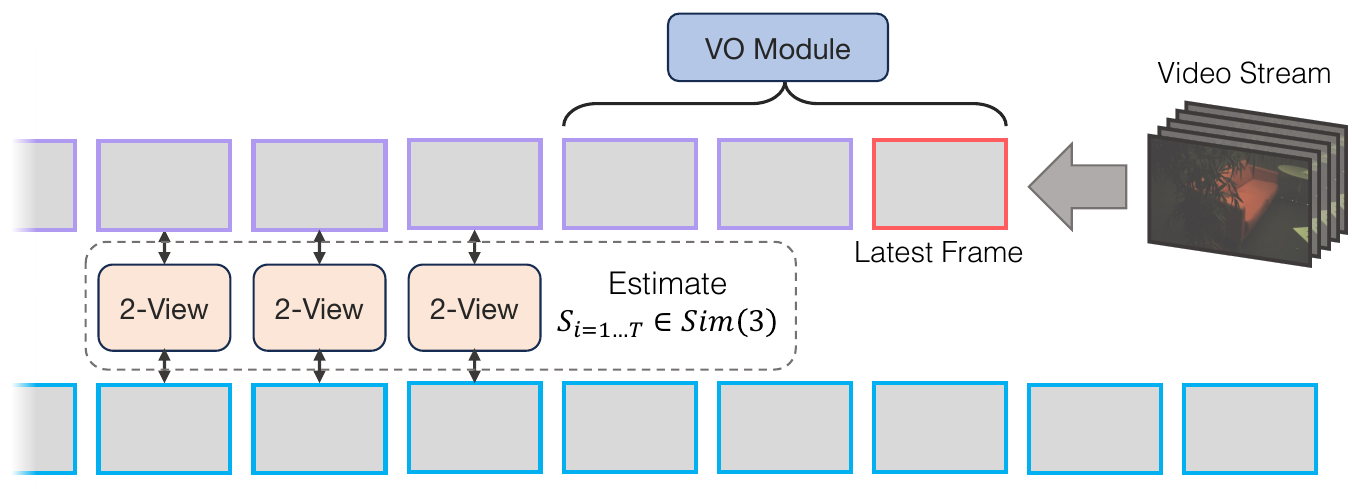}
    \caption{Our method estimates camera pose from multiple disconnected video streams.}
    \label{fig:framework}
\end{figure}
We propose a method for Multi-Session SLAM, capable of both wide-baseline relative pose and visual odometry using a single backbone architecture inspired by \cite{dpvo}. We introduce a differentiable solver layer which minimizes the symmetric epipolar distance (SED) from bi-directional optical flow. From this we construct a method for two-view pose which is capable of matching from far-apart views. This same design can be repurposed for visual odometry by swapping out the solver for bundle adjustment. By employing a unified backbone architecture for both tasks, we enable a simple approach to Multi-Session SLAM.

We evaluate our approach on challenging real-world datasets: EuRoC-MAV~\cite{burri2016euroc} and ETH3D~\cite{eth3d}. Our system is more accurate than prior approaches, and is robust to catastrophic failures. We also evaluate our two-view pose method in isolation on the Scannet and Megadepth datasets, and show that it is competitive with transformer-based matching networks. For pairs of far-apart views, our method is capable estimating accurate relative pose.

Our backbone predicts iterative updates to optical flow coupled with a differentiable solver layer for estimating camera pose. This framework, based on RAFT~\cite{teed2020raft}, has worked exceptionally well for Visual Odometry~\cite{dpvo} and SLAM~\cite{teed2021droid}. By leveraging this idea for wide-baseline matching, we can extend these methods to the Multi-Session setting without introducing substantial complexity.\looseness=-1

\section{Related Work\vspace{-2mm}}

\begin{figure*}[t]
    \centering
    \includegraphics[width=0.95\textwidth]{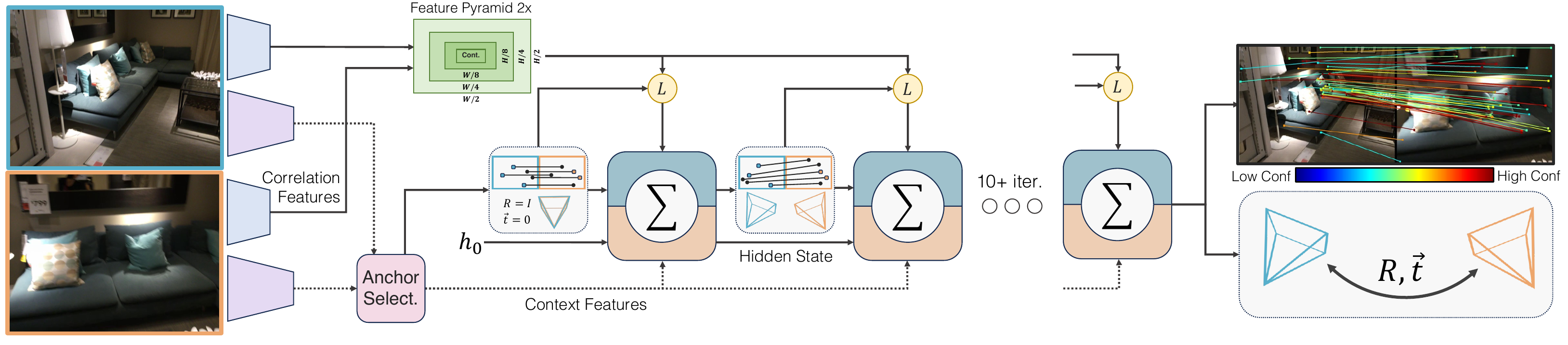}
    \caption{Overview of our backbone. Given a group of $\geq 2$ frames, our method jointly estimates bi-directional optical flow and camera poses. This module is applicable to (1) wide-baseline two-view matching and (2) visual odometry. We use our backbone for Multi-Session SLAM, which requires the ability to perform both (1) and (2). For each image, we select a set of 2D anchor points and initialize their depth and the camera poses trivially. Our approach then iteratively refines the matches for each anchor, similar to RAFT~\cite{teed2020raft}, while updating the camera poses. We alternate between matching and pose updates, where each one informs the update to the other. This entire procedure is repeated several times until convergence.\vspace{-3mm}}%
    \label{fig:twoview_update_op}
\end{figure*}

Most prior works treat Visual Simultaneous Localization and Mapping (Visual SLAM) as an optimization problem solving for a 3D scene model and camera trajectory which best explains the visual measurements~\cite{cadena2016past}.
\smallskip\\
\textbf{Indirect} approaches to SLAM perform keypoint matching as a pre-processing step, then estimate a 3D point cloud and camera poses by optimizing the 2D reprojection error over all matches~\cite{orbslam1, orbslam2, orbslam3, leutenegger2013keyframe, schoenberger2016sfm, sarlin2019coarse}. Reprojection-error is easier to optimize than photometric alignment, making indirect methods robust to lower camera hz~\cite{teed2021droid}. Keypoint matching also enables a straightfoward approach to estimating two-view relative pose, a necessary step for Visual Multi-Session SLAM~\cite{hartley1997defense, lepetit2009ep, lu2018review, umeyama1991least, sarlin2019coarse}. However, keypoint-based SLAM is less robust to low-texture environments compared to those which use photometric alignment~\cite{lsdslam, dso} or optical flow~\cite{teed2021droid,dpvo}.\looseness=-1 %
\smallskip\\
\textbf{Semi-Indirect} approaches similarly optimize 2D reprojection error like indirect methods, but without requiring matches as input~\cite{dpvo, teed2021droid, teed2018deepv2d}. Instead, these approaches alternate between predicting optical flow residuals and performing bundle adjustment. Semi-indirect methods do not require repeatable keypoints across images, making them robust to low-texture settings while retaining the easier reprojection-error objective.%

Our approach is most similar to Deep Patch Visual Odometry~\cite{dpvo} (DPVO), which is a sparse analog of DROID-SLAM. DPVO predicts sparse optical flow instead of dense, performing similarly to DROID-SLAM while running faster and using half the memory.%
\smallskip\\
\textbf{Differentiable solver layers} for camera pose estimation have been used in order to learn outlier rejection with data-driven training. For two-view relative pose, Ranftl and Koltun~\cite{deepfundmatrixest} used a deep network to learn an iteratively reweighted least-squares algorithm, which solved a weighted variant of the 8-point-algorithm~\cite{hartley1997defense} using confidences predicted by the network. \cite{deepfundmatrixest} required matched points as input, wheras our approach can work from images alone. Roessle and Nie{\ss}ner~\cite{roessle2023end2end} proposed an end-to-end architecture which used the weighted 8-point-algorithm from ~\cite{deepfundmatrixest} on top of matches produced using Superglue~\cite{sarlin2020superglue}, and supervised directly on the predicted pose. \cite{roessle2023end2end} works well, but is unable to outperform existing methods that use minimal solvers with LO-RANSAC~\cite{PoseLib} instead of differentiable solvers. In contrast, our approach is not built on top of an existing SOTA 2-view matcher; our design is also iterative, applying a recurrent module and solver layers multiple times to refine the prediction.
\smallskip\\
\textbf{Multi-Session SLAM} is the task of performing SLAM on multiple trajectories of the same scene. Like SLAM, Multi-Session SLAM is an online task where camera motion is estimated from a stream of images. However, in the Multi-Session setting, there are known breaks in the data-stream over which the small-baseline assumption no longer holds. While local optimization is sufficient for motion tracking from video, estimating wide-baseline camera pose is often non-convex~\cite{strasdat2010scale} and more challenging due to large viewpoint changes. In monocular visual SLAM, the scale of each sequence is also ambiguous; aligning two trajectories requires estimating 7 degrees-of-freedom (translation, rotation, scale) for all sequences, excluding the first which can be considered the reference.\looseness=-1

ORB-SLAM3~\cite{orbslam3} performs Multi-Session SLAM by matching between ORB~\cite{rublee2011orb} descriptors. While tracking camera motion over each sequence, ORB-SLAM3 continually updates an \emph{Atlas} of the scene - a lookup table between 2D keypoints and their 3D map point in the reference frame of their original sequence. To align disjoint sequences, candidate cross-sequence image pairs are identified using image-retrieval~\cite{dBoW}, the 3D map points are obtained from the Atlas, and the relative transformation is found using the Umeyama algorithm~\cite{umeyama1991least}. Several methods perform Multi-Session Visual Inertial SLAM~\cite{labbe2022multi, qin2018vins}, however we focus on the visual-only monocular setting where the scale of each session is ambiguous and must be estimated when joining sequences.\looseness=-1

CCM-SLAM~\cite{schmuck2017multi, schmuck2019ccm} and SLAMM~\cite{daoud2018slamm} perform Multi-Session visual SLAM and are built on top of ORB-SLAM~\cite{orbslam1} as well, but are optimized for limited bandwidth and distributed processing, while ORB-SLAM3 performs better and is optimized for accuracy and speed.\vspace{-3mm}

\section{Approach}

\noindent\textbf{Overview:} We propose a backbone for matching between ${\geq} 2$ views, and then construct a method for multi-session SLAM upon it. Our backbone approaches matching as optical flow; it borrows several ideas from RAFT~\cite{teed2020raft}, such as iteratively predicting flow residuals using a recurrent network, and the correlation feature pyramid. Our method\ maintains a running estimate of both camera pose and bi-directional optical flow, and uses the update operator to refine them both. We provide an overview in Fig.~\ref{fig:twoview_update_op}. An invariant of our backbone is that the matches are always clamped to plausible values given the pose estimates and an assumption of rigid scene geometry, therefore WLOG the matches can be considered a depth estimate.\looseness=-1

At initialization, our backbone produces dense correlation feature pyramids for each image and context features which remain fixed throughout the forward pass. The update operator uses an RNN to predict an update to the matches, and a differentiable solver to update poses. Our backbone treats the two-view and multi-view settings differently: In the two-view setting, the solver updates the poses to minimize the symmetric epipolar distance (SED). In the multi-view setting, the solver updates both poses and depth to minimize the reprojection error. After the solver layer, the matches are adjusted to agree with the poses/depth.%

\subsection{Initialization}
\begin{figure*}[t]
    \centering
    \includegraphics[width=0.95\textwidth]{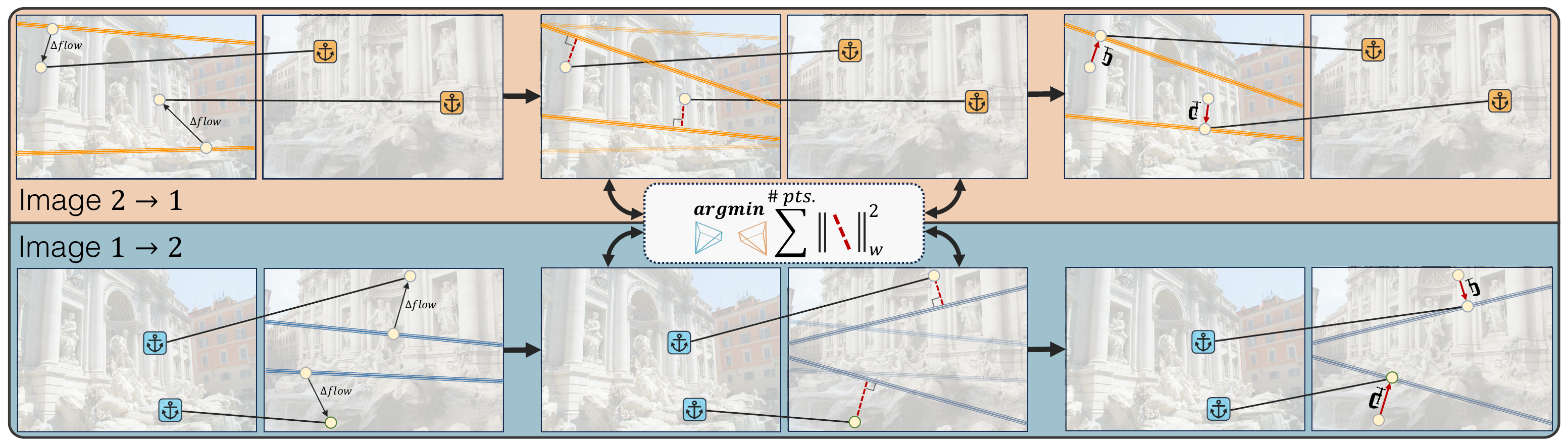}
    \caption{A single update iteration (two-view). For each anchor-match pair, we predict an update to the matches using the RNN. We then solve for an update to the camera poses which minimizes the symmetric epipolar distance (SED), producing a new set of epipolar lines. Finally, we clamp the matches back to the best-fit epipolar lines, and repeat the whole process again. In Fig.~\ref{fig:qual_iters} and the Appendix, we visualize these iterations on real-world images.\vspace{-3mm}}
    \label{fig:iterations}
\end{figure*}

\noindent\textbf{Feature Extraction and Feature Pyramid:} Similar to RAFT, our method\ separately extracts context and correlation feature for each image. The context features are provided as input to the recurrent update operator, whereas the correlation features are used to evaluate the visual similarity between any two pixels using a dot product. The context feature maps are produced at $\nicefrac{1}{8}$ resolution using a residual network. We then apply a linear-self-attention residual for long-range feature sharing. The correlation feature maps are also produced using a residual network, but with several exit ramps to produce features at $\nicefrac{1}{2}$, $\nicefrac{1}{4}$, and $\nicefrac{1}{8}$ of the input image resolution. We also average-pool the last one three additional times to produce a correlation feature pyramid with 6 levels. We depict their architecture in the Appendix.\looseness=-1

\smallskip\noindent\textbf{Anchor-Point Selection:} Our method predicts sparse optical flow, where matches have one end anchored and the other end free to move in $\mathbb{R}^2$. We use a mix of detector~\cite{detone2018superpoint}-chosen and randomly-chosen anchor points. Each point is assigned an initial match in the other image(s) at the same pixel coordinate. For the rest of the paper, we will refer to the $k^{th}$ anchor point as $a_k$, and its match in image $j$ as $m_{kj}$. 

We index the context feature map at each $a_k$ to produce a unique context feature vector $ctx_k \in \mathbb{R}^{384}$, which is used by the update operator. Beyond this point, the full context feature map is not used and is discarded to save memory.

\smallskip\noindent\textbf{Correlation:} Similar to other RAFT-based methods, we use correlation features to assess the visual alignment/similarity given by the current matching estimate. For each $(a_k,m_{kj})$ pair, we bilinearly sample the feature pyramids $f$ and $g$ for their respective frames at locations $a_k$ and $m_{kj}$ and take their inner product at each level, producing\vspace{-2mm}
\begin{equation}
\label{eq:corr0}
    \langle f_{k}^{i}, g_{kj}^i \rangle \in \mathbb{R} : i=1...6\\\vspace{-2mm}
\end{equation}
To provide additional spatial context, we also perturb both $a_k$ and $m_{kj}$ in $3\times 3$ and $7 \times 7$ grids, respectively, and calculate eq.~\ref{eq:corr0} for all pairs. The resulting feature vector $\mathbf{C}_{kj} \in \mathbb{R}^{(6 \times 3 \cross 3 \cross 7 \cross 7)}$ is then passed to the update operator. %

\subsection{Update Operator}
The update operator produces a revision to all $m_{kj}$ and the estimated relative pose. It consists of three stages, depicted in Fig.~\ref{fig:iterations}: (1) An RNN predicts a 2D update to $m_{kj}$ and an associated confidence weight $w_{kj}$. (2) We solve for a pose estimate which is consistent with the newly predicted matches and confidence. (3) We adjust the matches to be physically plausible assuming rigid scene geometry. In the two-view setting, we clamp the matches to the epipolar lines, and in the multi-view setting we reproject the anchor points using the depth and poses from the solver. %

Through the recurrent iterations, our method\ maintains a running state for each anchor-match $(a_k,m_{kj})$ pair, consisting of a hidden state vector $h_{kj}$, a confidence weight $w_{kj}$, and a match location $m_{kj}$ in frame $j$. This state is continually updated, during which $m_{kj}$ should ideally approach the true match location of $a_k$ in frame $j$, and $w_{kj}$ should approach 1. This is what we observe empirically.\vspace{-1mm}%

\begin{figure}[h]
    \centering
    \includegraphics[width=\columnwidth]{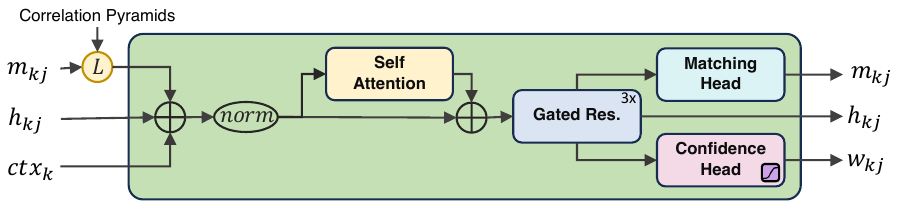}
    \caption{The RNN Module. For each anchor-match pair $(a_k, m_{kj})$, it predicts an update to $m_{kj}$ and an associated confidence $w_{kj}$. It also updates a hidden state $h_{kj} \in \mathbb{R}^{384}$. Internally, the RNN shares features via attention between updates with the same incoming and outgoing frame.\vspace{-3mm}}
    \label{fig:update_op}
\end{figure}

\smallskip\noindent\textbf{RNN Module:} The learnable component of the update operator is the recurrent network which predicts updates to $h_{kj}$, $m_{kj}$ and $w_{kj}$ for all anchor points. We visualize this operator in Fig.~\ref{fig:update_op}.

The input to the RNN are the context features, the previous hidden state, and the correlation features generated from the current matching estimate, all of which are added together and normalized with layernorm. A self-attention residual is also applied to edges with the same source and destination frame, followed by three gated-residual-units, whose architecture is depicted in the Appendix. The output is an updated hidden state, from which we predict an updated match and confidence using the flow head and confidence head, respectively, which are implemented as two-layer MLPs. The confidence head includes a softmax to restrict $w_{kj} \in (0,1)$.\vspace{-2mm}

\subsection{Two-View Solver and Pre-Conditioning}

\begin{figure}[t]
    \centering
    \includegraphics[width=0.95\columnwidth]{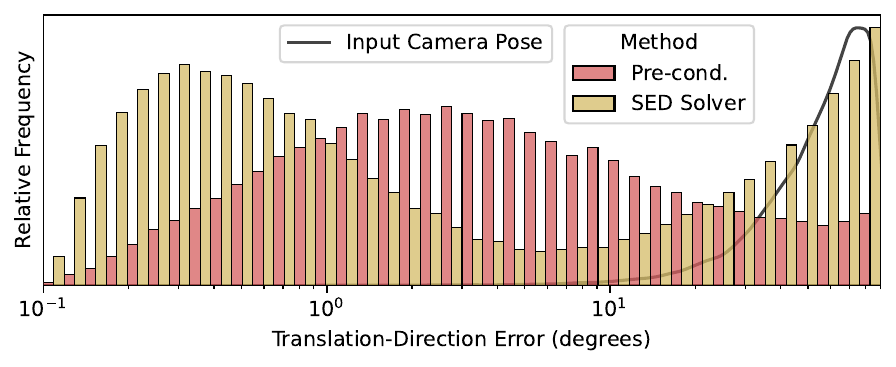}
    \caption{Convergence of our solver, given ground-truth matches (from TartanAir~\cite{wang2020tartanair}). Our SED solver is accurate, but only converges to the global minimum when initialized close to it. We initialize the pose using the 8-pt-algorithm to ensure it is within the SED convergence basin.\vspace{-3mm}}
    \label{fig:dir_error}
\end{figure}

Our two-view solver aims to align the relative pose between two frames to the predicted matches by minimizing the distance between each match and its epipolar line. This strategy is accurate, however the objective function is non-convex, meaning it will only converge to the global minimum if it is initialized close to it. In contrast, the 8-point-algorithm, a common approach to this task, has the \emph{opposite} problem; it does not suffer from local minimum since it is a homogeneous least squares, but the solution is sub-optimal. In Fig.~\ref{fig:dir_error}, we visualize how our solver either converges to within a fraction of a degree, or not at all, meanwhile the 8-point-algorithm is more robust but less accurate.

The approach our method uses is to pre-condition the pose estimate using a weighted, dense variant of the 8-point algorithm~\cite{deepfundmatrixest}, and then run our solver layer to refine the pose. This combined strategy obtains the best of both worlds, since the pre-conditioning will typically initialize the pose within the basin of convergence of our solver, which then refines the prediction. Fig.~\ref{fig:dir_error} visualizes this basin on the TartanAir~\cite{wang2020tartanair} dataset using ground-truth flow.

\smallskip\noindent\textbf{Pre-conditioning:} We solve the homogeneous least squares problem~\cite{hartley1997defense}\vspace{-1mm}
\begin{equation}
   \argmin_{\textbf{F}} ||diag(\vec{w})\mathbf{M}\vec{\textbf{F}}||^2 \quad s.t. \ ||\textbf{F}||^2 = 1
\vspace{-2mm}\end{equation}
where $\mathbf{M}$ and $\vec{w}$ are constructed from the anchor-match pairs and confidence predictions. The points are normalized to $[-1,1]$ beforehand. Afterwards, we reshape and uncalibrate $\mathbf{F}$ to obtain the essential matrix.

We obtain the four relative pose candidates following the procedure detailed in ~\cite{cvbook} and in the Appendix. During training, we select the pose candidate closest to the ground-truth on the $SE(3)$ manifold. During inference, we select the candidate by testing for chirality~\cite{cvbook}.\smallskip

\begin{figure}[t]
    \centering
    \includegraphics[width=0.95\columnwidth]{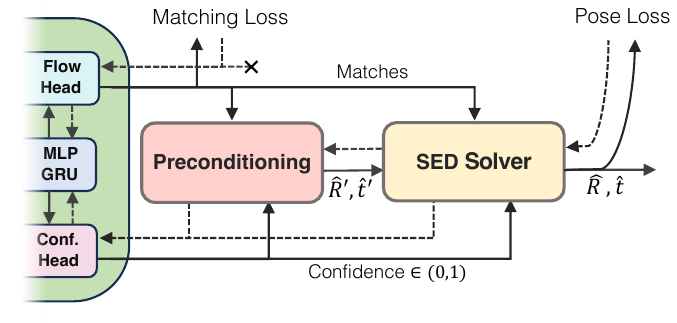}
    \caption{The flow of gradients through our solver. The updated matches from the RNN are supervised directly, and then detached from the gradient tape. The solver output is supervised with a pose loss. In the backward pass, gradients from the pose loss are used to supervise the confidence head in order to learn outlier rejection.\vspace{-3mm}}
    \label{fig:gradflow}
\end{figure}
\noindent\textbf{SED Solver Layer:} The solver layer seeks to minimize the distance between each match $m_{kj}$, and the epipolar line induced by the current pose estimate and its respective anchor point $a_k$. The solver objective is bi-directional, since every image contains a unique set of anchor points; the objective is also known as the titular symmetric epipolar distance (SED)~\cite{fathy2011fundamental}. We parameterize the pose update as small rotations to the translation direction and orientation: $\xi_{\mathbf{t}}, \xi_\mathbf{R} \in \mathfrak{so}(3)$. These are the free variables. Let $K(i)$ be the set of anchor indices for frame $i$, $\mathbf{R}$ and $\mathbf{t}$ be the output of the pre-conditioning stage, and $epi(x,R,t)$ compute the epipolar line for a point $x$ given the relative pose $(R,t)$. Formally, the minimization objective in our solver is
\begin{equation}\label{eq:obj}
\begin{gathered}
    l_{kj} = epi(a_k, (e^{\xi_\mathbf{R}} \mathbf{R})_{i\rightarrow j}, (e^{\xi_\mathbf{t}}\mathbf{t})_{i\rightarrow j}) : k\in K(i)\\
    E_{i\rightarrow j} = \sum_{k \in K(i)} w_{kj} \cdot \norm{err(m_{kj}, l_{kj})}_2^2\\
    SED = \argmin_{\xi_\mathbf{R}, \xi_\mathbf{t}} \Big(E_{i\rightarrow j} + E_{j\rightarrow i}\Big)
\end{gathered}
\end{equation}
where $err$ computes the 2D point-to-line error. The predicted weights $w_{kj}$ are included to allow the network to down-weight the contribution of any match which it deems unreliable. This solver layer is implemented in Pytorch, enabling gradients to propagate backward from the pose loss to the confidence head using autograd. We depict the gradient flow in Fig.~\ref{fig:gradflow}.

To minimize eq.~\ref{eq:obj}, we employ the Levenberg Marquardt algorithm. This requires computing the Jacobian for all terms. We provide the full derivations in the Appendix, but define the residual function $err$ in its entirety here. 

\smallskip\noindent Let $l$ be the epipolar line produced by $(a, e^{\xi_\mathbf{R}} \mathbf{R}, e^{\xi_\mathbf{t}}\mathbf{t})$. We can express each term of eq.~\ref{eq:obj} as:
\begin{equation}
    err(m,l) = \Bigg(\frac{l_xm_x + l_ym_y + l_z}{l_x^2 + l_y^2}\Bigg)\begin{bmatrix}
    l_x \\
    l_y
  \end{bmatrix}
\end{equation}
and the epipolar line as
\begin{equation}
\begin{gathered}
  \mathbf{E} = (e^{\xi_\mathbf{R}}\mathbf{R})^\top [ e^{\xi_\mathbf{t}}\mathbf{t}]_\cross \\
    l = \mathbf{F}\begin{bmatrix}
    a_x \\
    a_y \\
    1
  \end{bmatrix} = \Biggl[(K_2^\top)^{-1}\mathbf{E}K_1^\top\Biggr] \begin{bmatrix}
    a_x \\
    a_y \\
    1
  \end{bmatrix} 
\end{gathered}
\end{equation}
where $\mathbf{F}$ and $\mathbf{E}$ are the fundamental and essential matrices.%
\subsection{Adapting our backbone to Visual Odometry}

\noindent To adapt our backbone to VO, we make several changes:

\smallskip\noindent\textbf{Multi-view Solver (BA):} Our local optimizer minimizes reprojection error and treats depth as a separate variable. This is identical to the bundle adjustment from DPVO. Formally, let $F$ be the set of connected frames, $G$ be the global poses, $d_k$ be the depth estimate for anchor $k$, and $\Pi(\cdot)$ be the $3D\rightarrow 2D$ projection function. The bundle adjustment objective is:
\begin{equation}
\label{eq:bundleadjustment}
\argmin_{G, \mathbf{d}} \sum_{(i,j)\in F} \sum_{k\in K(i)}  w_{kj} \cdot \norm{\Pi[G_{j}^{-1} G_i \Pi^{-1}(a_k,d_k)] - m_{kj}}^2
\end{equation}
The preconditioning stage is not necessary as we can linearly-extrapolate the camera pose estimates from previous frames to achieve good initialization.

\smallskip\noindent\textbf{Clamping:} Since epipolar lines do not make sense in the multi-view setting, we reset the optical flow using the pose and depth from the bundle adjustment. 

\smallskip\noindent\textbf{RNN changes:} We include mechanisms in the RNN to share latent features between updates if they stem from the same anchor point. Specifically, we use the message passing and temporal convolutions from DPVO~\cite{dpvo}.
\subsection{Trajectory Alignment}\label{sec:sim3est} In the Multi-Session SLAM problem, we estimate a relative $\Simt$ between pairs of disjoint trajectories in order to align them. All anchor points already have a depth estimate $d_k$ as result of the bundle adjustment (eq.~\ref{eq:bundleadjustment}) from the VO system. We use our two-view method to estimate the relative rotation and translation direction, and compute the translation magnitude and relative scaling by comparing the $d_k$'s to the depth from the triangulated two-view matches. %
\smallskip\\
\textit{1) Estimate relative rotation, translation direction, and matches}. We first retrieve a candidate image pair $(i, j)$ using NetVLAD~\cite{netvlad}, one from each trajectory, and apply our two-view model to estimate their relative rotation and translation direction.%
\smallskip\\
\textit{2) Align the depth from the two-view and VO operators}. We previously defined $d_k$ as the depth output of the VO system. Let $d^{\prime}_k$ be the triangulated depth obtained from our two-view matches:\vspace{-2mm}%
\begin{equation}
    \label{eq:depthj}
    d^{\prime}_{k} = \begin{cases} 
    triang(\mathbf{R}_{i\rightarrow j}, \mathbf{t}_{i\rightarrow j}, a_k, m_{kj}) & : k \in K(i)\\
    triang(\mathbf{R}_{j\rightarrow i}, \mathbf{t}_{j\rightarrow i}a_k, m_{ki}) & : k \in K(j)\\
\end{cases}
\end{equation}
We solve\vspace{-4mm}
\begin{equation}
    \label{eq:ransac_mag}
    \argmax_{s_{i} \in \mathbb{R} > 0} \sum_{k\in K(i)} \mathbf{1} \Bigg[\frac{1}{\lambda} < \frac{d_{k}}{s_i\cdot d^{\prime}_{k}} < \lambda \Bigg]
    \vspace{-2mm}
\end{equation}
to recover the translation magnitude, where the hyperparameter $\lambda = 1.05$. In layman terms, eq.~\ref{eq:ransac_mag} seeks \emph{what translation magnitude would align the most triangulated points to the existing 3D map?} We can obtain a reasonably good solution to eq~\ref{eq:ransac_mag} by brute-force checking $s_i=d_{k} / d_{k}^{\prime} \quad \forall k\in K(i)$, similar to RANSAC. If the number of inliers is too small, we retry on a new candidate pair. Conversely, the scale-difference is $s_i / s_j$, where $s_j$ is estimated the same way, but for anchor points in $K(j)$:\vspace{-3mm}
\begin{equation}
    \label{eq:ransac_mag}
    \argmax_{s_j \in \mathbb{R} > 0} \sum_{k \in K(j)} \mathbf{1} \Bigg[\frac{1}{\lambda} < \frac{d_{k}}{s_j\cdot d_{k}^{\prime}} < \lambda \Bigg]
\end{equation}
Given the rotation $\mathbf{R}_{j\rightarrow i}$ and translation $\mathbf{t}_{j\rightarrow i}$ from the two-view solver layers, the resulting transformation can be used to align the two trajectories:\vspace{-2mm}
\begin{equation}
      \mathcal{S}_{j\rightarrow i} = \begin{bmatrix}
    \frac{s_{i}}{s_j}\mathbf{R} & s_{j}\mathbf{t} \\
    0 & 1
  \end{bmatrix} \in \Simt
  \vspace{-2mm}
\end{equation}
The graphs are then merged by concatenating all buffers.%

\subsection{Training}

We train our backbone separately for VO and for two-view pose. It is trained using a matching loss and a pose loss. The pose loss for the two-view training
\begin{equation}
    \mathcal{L}_{pose2V} = \sum_{t=1}^{12} cos^{-1}\bigg(\bar{\textbf{t}^t} \cdot \bar{\textbf{t}^{\textbf{gt}}}\bigg) + \alpha \norm{[(\textbf{R}^t)^T \textbf{R}^{\textbf{gt}}]}_{SO(3)}
    \vspace{-1mm}
\end{equation}
penalizes angle error for both the predicted translation direction and predicted orientation. The pose loss for the VO training is the SE3 manifold distance between the predicted and ground-truth poses:\vspace{-3mm}
\begin{equation}
    \mathcal{L}_{poseVO} = \sum_{t=1}^{T} \norm{[(\textbf{G}^t)^{-1} \textbf{G}^{\textbf{gt}}]}_{SE(3)}
    \vspace{-2mm}
\end{equation}
The matching loss\vspace{-3mm}
\begin{equation}
    \mathcal{L}_{matching} = \frac{1}{|N|}\sum_a^N\sum_{t=1}^{T} \norm{m_a^t - m_a^{\textbf{gt}}}_2
    \vspace{-2mm}
\end{equation}
is a standard endpoint-error~\cite{teed2020raft,lipson2021raft}, applied only on pixels which have valid depth and are verified to be visible in both images. Our update operator is applied 12 times per training example; supervision is applied after every update (See Fig.~\ref{fig:gradflow}). The final loss is $\mathcal{L}=\mathcal{L}_{pose} + \beta \cdot \mathcal{L}_{matching}$.

Following prior work~\cite{lindenberger2023lightglue}, we pre-train our two-view method on synthetic homographies (without pose loss) for two epochs on the Oxford-Paris 1M Distractors dataset~\cite{RITAC18}. We then tune our model on a $\nicefrac{50}{50}$ mixture of Scannet~\cite{scannet} and Megadepth~\cite{MegaDepthLi18} for 100,000 steps and a batch size of 120, using 10 A6000 GPUs for 5 days. The VO backbone is trained using the procedure from \cite{dpvo} on TartanAir.\vspace{-1mm}

\begin{figure}[t]
    \centering
    \includegraphics[width=\columnwidth]{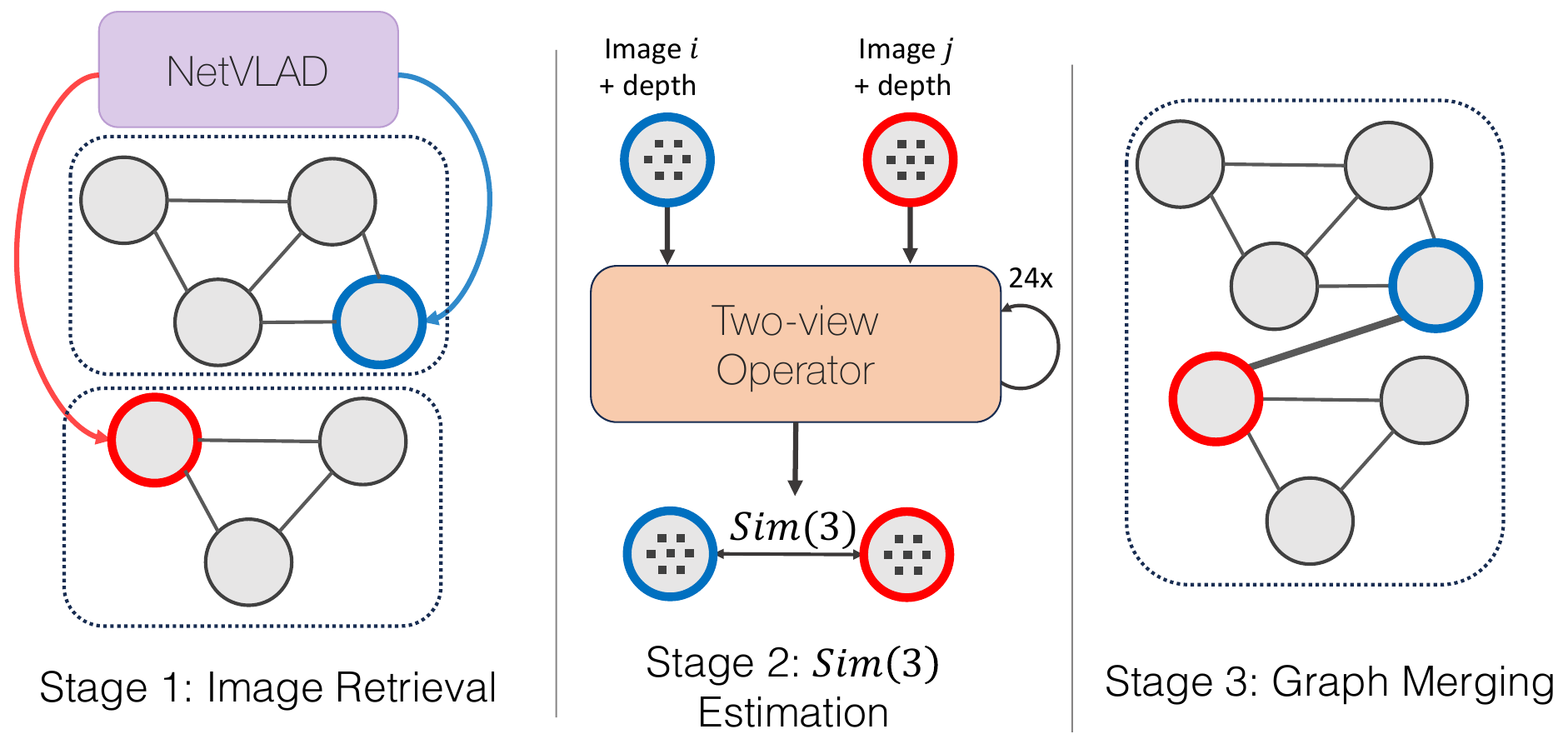}
    \caption{High-level overview of our approach to aligning disjoint trajectories. We use NetVLAD~\cite{netvlad} to retrieve two co-visible images, one from each trajectory, and apply our two-view update operator to estimate a relative $\Simt$ alignment, following the procedure detailed in Sec~\ref{sec:sim3est}. We then transform the second trajectory into the reference frame of the first, and merge all buffers.\vspace{-2mm}}
    \label{fig:sim3_est_fig}
\end{figure}

\subsection{Multi-Session SLAM System\vspace{-0mm}}

\noindent\textbf{Overview:} To perform Multi-Session SLAM, we use our VO-trained backbone to perform visual odometry and global optimization; our two-view backbone connects disjoint trajectories following the procedure in Sec~\ref{sec:sim3est}. Our full system builds the pose graph incrementally: new video frames inserted into the factor graph as they are received, and unjoined trajectories are aligned and merged as soon as a connection is found.%
\smallskip\\
\textbf{VO Frontend:} Our VO frontend extracts features and estimates camera poses and depth for incoming frames, operating on a sliding window covering the most recently observed 22 keyframes. Keyframing occurs retroactively on the $4^{th}$ oldest keyframe if it is found to be redundant. New poses are initialized using a linear-motion model, and depth is copied from the previous frame. Like DPVO~\cite{dpvo}, we initialize after observing 8 frames with significant motion.%
\smallskip\\
\textbf{Global Optimization:} We perform global optimization by introducing proximity factors using the existing pose and depth measurements, following the approach from DROID-SLAM~\cite{teed2021droid}, and update the matches and poses/depth using the VO backbone. We only run the backend periodically, and after joining trajectories.
\smallskip\\
\textbf{Trajectory Joining:} We query the database of NetVLAD descriptors to find co-visible frames. If several pairs are retrieved, we compute relative poses in a mini-batch and use the estimate with the highest inlier-ratio.

\section{Experiments}

We evaluate the performance of our two-view method in isolation, and our Multi-Session SLAM system as a whole.\vspace{-2mm}

\subsection{Two-View Evaluation\vspace{-2mm}}

\begin{table}[tb]
\begin{center}
\resizebox{\linewidth}{!}{
  \begin{tabular}{l >{\centering\arraybackslash}p{3.35cm} >{\centering\arraybackslash}p{1.2cm} >{\centering\arraybackslash}p{1.2cm} >{\centering\arraybackslash}p{1.2cm}}
    \toprule
    &\multirow{2}{\linewidth}{\centering Overall Approach}& \multicolumn{3}{c}{Pose error AUC [\%] $\uparrow$} \\
    \cmidrule(lr){3-5}
    && @5\degree & @10\degree & @20\degree \\
    \midrule
    Superglue~\cite{sarlin2020superglue} & \multirow{2}{*}{Matching} & 16.2(17.7) & 33.8(35.6) & 51.8(54.2) \\
    LightGlue~\cite{lindenberger2023lightglue} & \multirow{2}{*}{$\downarrow$} & 16.5(19.4) & 33.4(36.9) & 50.1(53.5) \\
    LoFTR~\cite{sun2021loftr} & \multirow{2}{*}{RANSAC~\cite{itseez2015opencv}} & 22.1(25.7) & 40.1(45.0) & 47.6(61.4) \\
    MatchFormer~\cite{wang2022matchformer} & \multirow{2}{*}{(LO-RANSAC~\cite{PoseLib})} & 24.3(27.3) & 43.9(47.6) & 61.4(64.9) \\ 
    ASpanFormer~\cite{chen2022aspanformer} & & 25.6(28.4) & 46.0(48.8) & 63.3(65.8) \\
    \midrule
    Roessle{\&}Nie{\ss}ner~\cite{roessle2023end2end} & Matching$\rightarrow$W8PA & 20.7 & 41.6 & 61.7 \\
    \midrule
    Roessle{\&}Nie{\ss}ner~\cite{roessle2023end2end} & Matching$\rightarrow$W8PA$\rightarrow$BA & 25.7 & 47.2 & 66.4 \\
    \midrule
     & Optical-Flow &  &  &  \\
    Ours & $\myarrow[45]$\hspace{12mm}$\myarrow[315]$ & \textbf{30.5} & \textbf{50.9} & \textbf{67.5} \\
     & Clamp $\leftarrow$ Solver  &  &  &  \\
    \bottomrule
    \vspace{-10mm}
  \end{tabular}
  }
\end{center}
\caption{Two-view results on Scannet~\cite{scannet}. Existing methods perform matching as pre-processing, then use a solver; our method is iterative. Results in parenthesis use the LO-RANSAC~\cite{PoseLib} estimator. ``W8PA$\rightarrow$BA" = weighted 8-point-algorithm and bundle-adjustment. Our design outperforms prior approaches. We use the same model weights as in Tab.~\ref{tab:two_view_pose_megadepth}.\vspace{-0mm}}
\label{tab:two_view_pose_scannet}
\end{table}

\begin{table}[tb]
\begin{center}
\resizebox{\linewidth}{!}{
  \begin{tabular}{l >{\centering\arraybackslash}p{3.35cm} >{\centering\arraybackslash}p{1.2cm} >{\centering\arraybackslash}p{1.2cm} >{\centering\arraybackslash}p{1.2cm}}
    \toprule
    &\multirow{2}{\linewidth}{\centering Overall Approach}& \multicolumn{3}{c}{Pose error AUC [\%] $\uparrow$} \\
    \cmidrule(lr){3-5}
    && @5\degree & @10\degree & @20\degree \\
    \midrule
    Superglue~\cite{sarlin2020superglue} & \multirow{2}{*}{Matching} & 49.7(65.8) & 67.1(78.7) & 80.6(87.5) \\
    LightGlue~\cite{lindenberger2023lightglue} & \multirow{2}{*}{$\downarrow$} & 49.9(66.7) & 67.0(79.3) & 80.1(87.9) \\
    LoFTR~\cite{sun2021loftr} & \multirow{2}{*}{RANSAC~\cite{itseez2015opencv}} & 52.8(66.4) & 69.2(78.6) & 81.2(86.5) \\
    MatchFormer~\cite{wang2022matchformer} & \multirow{2}{*}{(LO-RANSAC~\cite{PoseLib})} & 53.3(66.5) & 69.7(78.9) & 81.8(87.5) \\
    ASpanFormer~\cite{chen2022aspanformer} & & 55.3(\textbf{69.4}) & 71.5(\textbf{81.1}) & 83.1(\textbf{88.9}) \\
    \midrule
    Roessle{\&}Nie{\ss}ner~\cite{roessle2023end2end} & Matching$\rightarrow$W8PA & 46.9 & 62.8 & 76.3 \\
    \midrule
    Roessle{\&}Nie{\ss}ner~\cite{roessle2023end2end} & Matching$\rightarrow$W8PA$\rightarrow$BA & 61.2 & 74.9 & 85.0 \\
    \midrule
     & Optical-Flow &  &  &  \\
    Ours & $\myarrow[45]$\hspace{12mm}$\myarrow[315]$ & 60.2 & 72.3 & 81.0 \\
     & Clamp $\leftarrow$ Solver  &  &  &  \\
    \bottomrule
    \vspace{-10mm}
  \end{tabular}
  }
\end{center}
\caption{Two-view results on Megadepth~\cite{MegaDepthLi18}. Existing methods perform matching as pre-processing, then use a solver; our method is iterative. ``W8PA$\rightarrow$BA" = weighted 8-point-algorithm and bundle-adjustment. Results in parenthesis use the LO-RANSAC~\cite{PoseLib} estimator. Our radically different design leads to better results indoors (Tab.~\ref{tab:two_view_pose_scannet}), but is less competitive in photo-tourism settings where high-volume matching is easier and therefore RANSAC-based methods fare better. We use the same model weights as in Tab.~\ref{tab:two_view_pose_scannet}. \vspace{-2mm}}
\label{tab:two_view_pose_megadepth}
\end{table}

We evaluate our two-view method on two popular relative pose benchmarks: The Scannet~\cite{scannet} 1500 and MegaDepth~\cite{MegaDepthLi18} 1500 test datasets. We compare to existing matching networks~\cite{sarlin2020superglue, lindenberger2023lightglue, sun2021loftr, wang2022matchformer, chen2022aspanformer}, which are typically in service of improving COLMAP~\cite{schoenberger2016sfm}-based SfM pipelines~\cite{sarlin2019coarse, lindenberger2021pixel}. In contrast, our approach is in service of Multi-Session SLAM. We report pose error AUC, as is done in prior matching work (higher is better). We use the same configuration and model weights on both datasets.

Most existing methods perform matching as a pre-processing step using a deep network, then estimate pose with a minimal solver using RANSAC~\cite{itseez2015opencv, PoseLib}. Roessle and Nie{\ss}ner~\cite{roessle2023end2end} replace the RANSAC optimizer with a weighted-8-point algorithm and bundle adjustment. In contrast, our two-view method couples the prediction of pose and matches, refining both over many iterations, and regresses optical flow instead of keypoint affinities.
\smallskip\\
\textbf{Scannet:} We report two-view results on Scannet~\cite{scannet} in Tab.~\ref{tab:two_view_pose_scannet}. Our approach outperforms existing methods on Scannet, which contains fewer salient keypoints and many texture-less surfaces and motion blur.%
\smallskip\\
\textbf{MegaDepth:} We report two-view results on Megadepth~\cite{MegaDepthLi18} in Tab.~\ref{tab:two_view_pose_megadepth}. On Megadepth, the matching problem is substantially easier since photo-tourism images contain many salient keypoints, resulting in better performance across all methods; our approach is on-par with existing approaches using the OpenCV RANSAC pose estimator~\cite{itseez2015opencv}, but not the LO-RANSAC~\cite{PoseLib} estimator.%\vspace{-1mm}
\smallskip\\
\textbf{Ablations:} We ablate on various aspects of our proposed architecture in Tab.~\ref{tab:ablations} and show that they lead to improved performance. Specifically, we show that pre-conditioning and the SED solver lead to better accuracy when used together, and that clamping the matches to the epipolar lines is also important.
\subsection{Multi-Session SLAM evaluation\vspace{-1mm}}
\label{sec:multi_sess_slam}
\begin{figure*}[t]
    \centering
    \includegraphics[width=\textwidth]{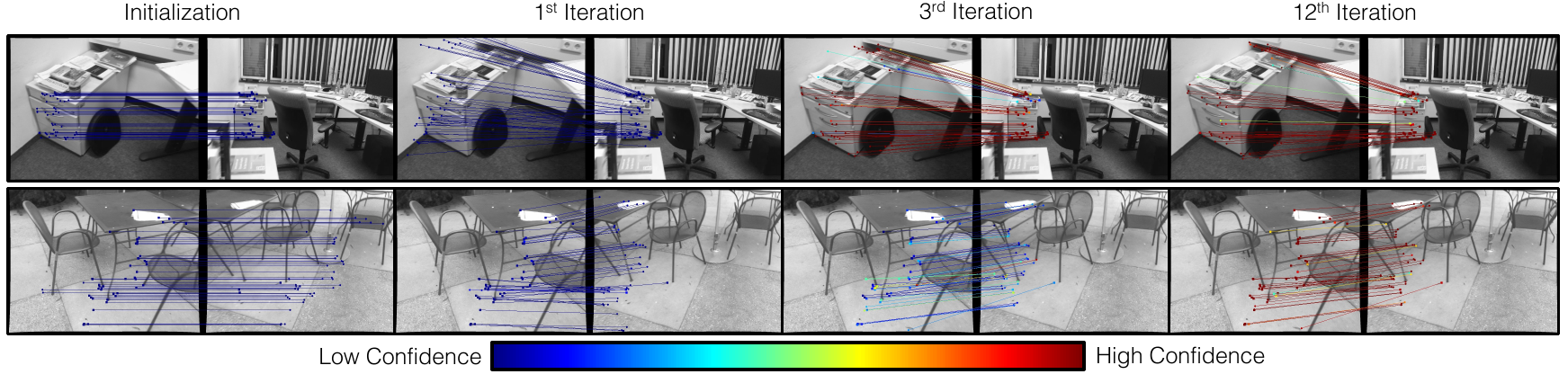}
    \caption{Qualitative results on Scannet. Our two-view variant is able to estimate accurate relative poses across wide camera baselines. It initializes all matches with uniform depth and identity relative pose. Progressive applications of our update operator lead to more accurate matches and higher predicted confidence. Some image pairs (row 2) take more iterations to converge than others (row 1).}
    \label{fig:qual_iters}
\end{figure*}

\begin{figure}[t]
\begin{center}
 \begin{subfigure}{\columnwidth}
\resizebox{\linewidth}{!}{
  \begin{tabular}{ccccc|c}
    \toprule
    \multicolumn{1}{c}{Scene name} & MH01-03 & MH01-05 & V101-103 & V201-203 & Cam.\\
    \cmidrule(lr){1-1}\cmidrule(lr){2-2}\cmidrule(lr){3-3}\cmidrule(lr){4-4}\cmidrule(lr){5-5}
    \multicolumn{1}{c}{\# Disjoint Trajectories} & 3 & 5 & 3 & 3 & hz\\
    \midrule
    \midrule
    Ours & \multirow{2}{*}{\textbf{0.022}} & \multirow{2}{*}{\textbf{0.036}} & \multirow{2}{*}{\textbf{0.031}} & \multirow{2}{*}{\textbf{0.024}} & \multirow{2}{*}{20} \\
    Mono-Visual & & & & \\
    \midrule
    CCM-SLAM~\cite{schmuck2019ccm} & \multirow{2}{*}{0.077} & \multirow{2}{*}{-} & \multirow{2}{*}{-} & \multirow{2}{*}{-} & \multirow{2}{*}{\textbf{38}}\\
    Mono-Visual & & & & \\
    \midrule
    ORB-SLAM3~\cite{orbslam3} & \multirow{2}{*}{{0.030}} & \multirow{2}{*}{{0.058}} & \multirow{2}{*}{0.058} & \multirow{2}{*}{0.284} & \multirow{2}{*}{20} \\
    Mono-Visual & & & & \\
    \midrule
    VINS~\cite{qin2018vins} & \multirow{2}{*}{-} & \multirow{2}{*}{0.210} & \multirow{2}{*}{-} & \multirow{2}{*}{-} & \multirow{2}{*}{-} \\
    Mono-Inertial &&&&\\
    \midrule
    ORB-SLAM3 & \multirow{2}{*}{0.037} & \multirow{2}{*}{0.065} & \multirow{2}{*}{0.040} & \multirow{2}{*}{0.048} & \multirow{2}{*}{20} \\
    Mono-Inertial &&&&\\
    \bottomrule
    \vspace{-10mm}
  \end{tabular}
  }
  \end{subfigure}
\end{center}
\caption{Multi-Session SLAM evaluation on EuRoC-MAV~\cite{burri2016euroc} using RMSE ATE[m] $\downarrow$. We compare to existing visual and inertial Multi-Session SLAM approaches, all using monocular video input. Our method outperforms existing approaches on all sequence groups. Results for baseline approaches were obtained from \cite{orbslam3}.\vspace{-2mm}}
\label{fig:euroc_results}
\end{figure}

We evaluate our approach to Multi-Session SLAM on the EuRoC-MAV~\cite{burri2016euroc} and ETH3D~\cite{eth3d} datasets, since they provide camera poses under a single global reference. Following the evaluation for single-video SLAM, we align the final predictions to the ground truth by computing a global 7-DOF alignment to account for the guage freedoms. We report RMSE ATE[m] in metric units. We sample 96 anchors in each video frame using a mix of random and Superpoint~\cite{detone2018superpoint} keypoints. Our two-view method uses those same anchors to join pairs of trajectories. We use the same weights and configuration on both datasets.
\smallskip\\
\begin{figure}[t]
    \centering
    \includegraphics[width=0.95\columnwidth]{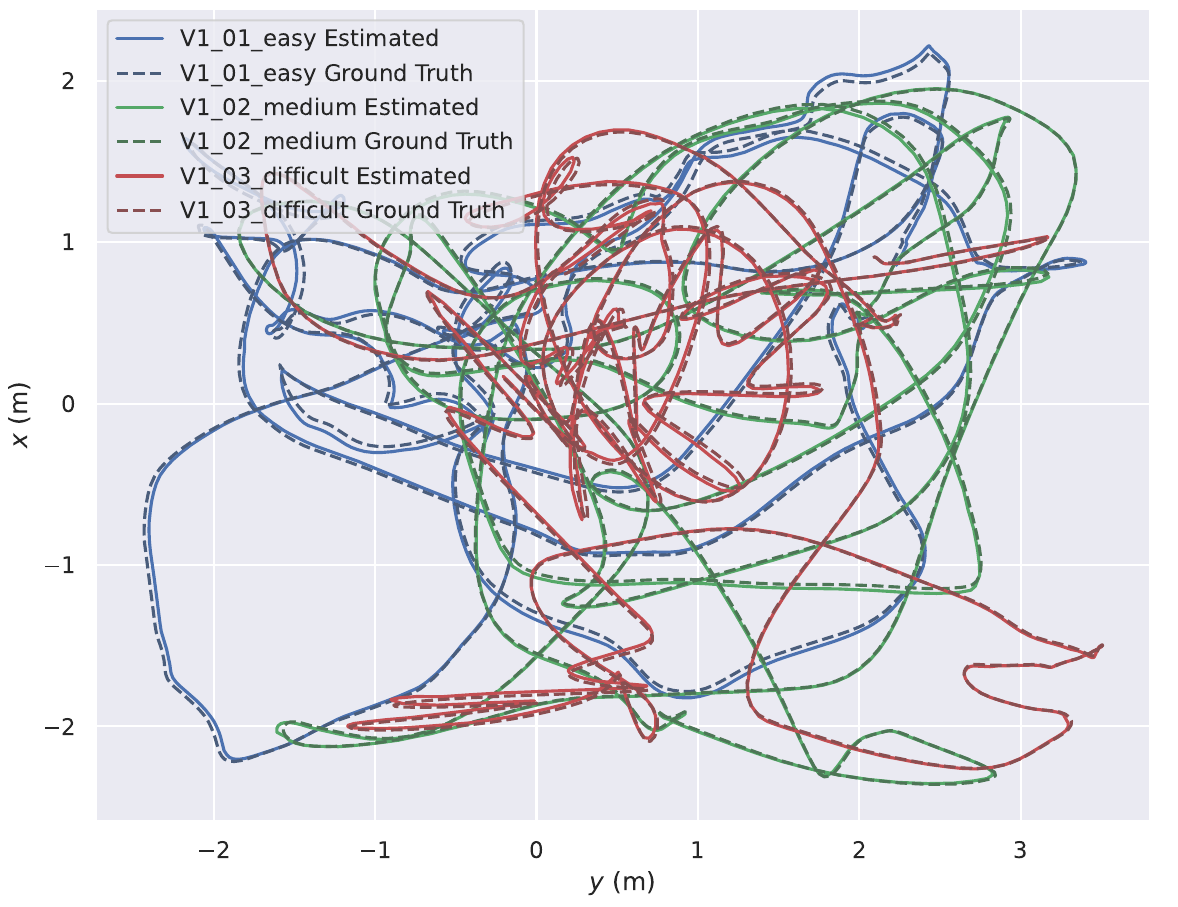}\vspace{-2mm}
    \caption{Our prediction on the Vicon1~\cite{burri2016euroc} sequences. Our method is accurate and robust to chaotic motion.\vspace{-3mm}}%
    \label{fig:v1_traj}
\end{figure}
\noindent\textbf{EuRoC-MAV:} In Tab.~\ref{fig:euroc_results} and Fig.~\ref{fig:v1_traj}, we report results on the EuRoC MAV dataset~\cite{burri2016euroc}. Following the evaluation from ~\cite{orbslam3}, we compare on four groups of disjoint trajectories across the three distinct environments: 3 in the Machine Hall, 5 in the Machine Hall, 3 in Vicon 1, and 3 in Vicon 2. Ground-truth poses are obtained using a laser tracker and Vicon cameras. These video sequences are long, with the entirety of the Machine-Hall group being 11.5 minutes of video (13.7K frames) spread across 5 sequences. Our approach achieves significantly lower error than ORB-SLAM3 on all groups, including 11x lower error on the Vicon 2 sequences (0.024 vs 0.284). Video is recorded at 20-FPS; all methods run in real-time. We also compare to CCM-SLAM~\cite{schmuck2019ccm}, which only provides results on the on MH01-03 sequence, the mono-inertial ORB-SLAM3, and the mono-inertial VINS~\cite{qin2018vins}. We outperform the other approaches on all trajectory groups.
\smallskip\\
\textbf{ETH3D:} In Tab.~\ref{tab:eth3d}, we report results on the training set from ETH3D~\cite{eth3d}. We compare across 5 unique scenes composed of multiple trajectories: 1-4 from \textit{Sofa}, 3\&4 from \textit{Table}, 1-3 from \textit{Plant Scene}, 1\&2 from \textit{Einstein}, and 2\&3 from \textit{Planar}. Sequences are excluded if they belong to the official ETH3D test set (with no ground-truth) or are from a different scene altogether. The sequences are trimmed from a single, longer video; to make joining the trajectories non-trivial, we reverse every other sequence, ensuring that there is a sufficient disconnect between subsequent videos. This is a novel benchmark, so \cite{schmuck2019ccm,qin2018vins} do not report results. \looseness=-1

ORB-SLAM3 fails on \textit{Sofa}, \textit{Plant Scene}, and \textit{Einstein} for various reasons (couldn't initialize, or lost the feature tracks). Our approach outperforms ORB-SLAM3 on 4/5 groups, and does not fail on any. ORB-SLAM3 only succeeds on 2/5 groups.%

\begin{table}[tb]
\begin{center}
\resizebox{\linewidth}{!}{
  \begin{tabular}{llcc}
    \toprule
    &\multirow{2}{*}{Ablation Experiment} & \multicolumn{2}{c}{Pose error AUC @10\degree[\%] $\uparrow$} \\
    \cmidrule(lr){3-4}
    && Megadepth Val & Scannet Val \\
    \midrule
    Baseline&Full model & \textbf{53.3} & \textbf{33.9} \\
    \midrule
    \multirow{3}{*}{Architecure}&Shallower correlation pyramid & 44.3 & 27.1 \\
    &No attention in update-op & 48.6 & 30.5 \\
    &No ReLU-attn in feat-extractor & 47.4 & 29.7 \\
    \midrule
    \multirow{3}{*}{Solver} & No pre-conditioning & 35.5 & 20.6 \\
     & No SED solver & 23.8 & 13.1 \\
    & No clamping step & 21.9 & 10.7 \\
    \bottomrule
    \vspace{-10mm}
  \end{tabular}
  }
\end{center}
\caption{Two-view ablation experiments on two validation sets. Our proposed RNN and feature extractor changes from DPVO improve the result. Removing components from our proposed solver design decreases accuracy, as does removing the clamping step.\looseness=-1\vspace{-1mm}}
\label{tab:ablations}
\end{table}

\begin{table}[tb]
\begin{center}
\resizebox{\linewidth}{!}{
  \begin{tabular}{cccccc}
    \toprule
    \multicolumn{1}{c}{Scene name} & Sofa & Table & Plant Scene & Einstein & Planar \\
    \cmidrule(lr){1-1}\cmidrule(lr){2-2}\cmidrule(lr){3-3}\cmidrule(lr){4-4}\cmidrule(lr){5-5}\cmidrule(lr){6-6}
    \multicolumn{1}{c}{\# Disjoint Trajectories} & 4 & 2 & 3 & 2 & 2 \\
    \midrule
    \midrule
    ORB-SLAM3~\cite{orbslam3} & \red{FAIL} & \multirow{2}{*}{{0.018}} & \red{FAIL} & \red{FAIL} & \multirow{2}{*}{\textbf{0.010}} \\
    Mono-Visual &\red{(no init)} & & \red{(init$\rightarrow$lost)} & \red{(init$\rightarrow$lost)} & \\
    \midrule
    Ours & \multirow{2}{*}{\textbf{0.010}} & \multirow{2}{*}{\textbf{0.010}} & \multirow{2}{*}{\textbf{0.021}} & \multirow{2}{*}{\textbf{0.032}} & \multirow{2}{*}{0.047} \\
    Mono-Visual &&&&\\
    \bottomrule
    \vspace{-10mm}
  \end{tabular}
  }
\end{center}
\caption{Multi-Session SLAM evaluation on ETH3D~\cite{eth3d} using RMSE ATE[m] $\downarrow$. Our method is robust to catastrophic failures.\vspace{-4mm}}
\label{tab:eth3d}
\end{table}

\section{Conclusion\vspace{-1mm}}

We introduce a new method for mono-visual Multi-Session SLAM. Our system utilizes a novel backbone which can estimate two-view pose and perform visual odometry. We leverage a novel differentiable solver which minimizes the symmetric epipolar distance. We compare against existing approaches and show strong performance across several datasets. This work was partially supported by IARPA and the National Science Foundation.

{
    \small
    \bibliographystyle{ieeenat_fullname}
    \bibliography{main}
}

\clearpage
\setcounter{page}{1}
\onecolumn
\begin{center}
{\Large \textbf{Appendix}}\\
\end{center}
\appendix

\section{Gradient Computation for the SED Solver}
\begin{align}
    d &\coloneq l_x^2 + l_y^2\\ 
    \zeta &\coloneq l_xm_x + l_ym_y + l_z\\
    \frac{\partial err(m,l)}{\partial l} &= \begin{bmatrix}
        \Big(\frac{-2l_x^2\zeta}{d^2} + \frac{l_xm_x}{d} + \frac{\zeta}{d}\Big) & \Big(\frac{-2l_xl_y\zeta}{d^2} + \frac{l_xm_y}{d}\Big) & \Big(\frac{l_x}{d}\Big) \\
        \Big(\frac{-2l_yl_x\zeta}{d^2} + \frac{l_ym_x}{d}\Big) & \Big(\frac{-2l_y^2\zeta}{d^2} + \frac{l_ym_y}{d} + \frac{\zeta}{d}\Big) & \Big(\frac{l_y}{d}\Big) \\
    \end{bmatrix}
\end{align}
\begin{center}
    Let $\Bar{a}$ be the anchor location in homogenous coordinates. The partials for $l$ w.r.t. $\mathbf{F}$ are
\end{center}
\begin{align}
    \frac{\partial l}{\partial \mathbf{F}} = \frac{\partial \mathbf{F} \Bar{a}}{\partial \mathbf{F}} = \Bar{a}^\top \otimes \mathbb{I} \in \mathbb{R}^{3 \times (3 \times 3)}
\end{align}
\begin{center}
    The partials for the fundamental matrix w.r.t. the essential matrix, given the known calibration matrix $K$:
\end{center}
\begin{align}
   \frac{\partial \mathbf{F}}{\partial \mathbf{E}} = \frac{\partial \Big[(K^{-1})^\top \mathbf{E} K^{-1}\Big]}{\partial \mathbf{E}} = (K^{-1})^\top \otimes K^{-1} \in \mathbb{R}^{(3 \times 3) \times (3 \times 3)}
\end{align}
\begin{center}
The $3 \times 3$ essential matrix in terms of the input rotation $\mathbf{R}$ and translation ${\mathbf{t}}$, and the local updates $\xi_{\mathbf{R}}$ and $\xi_{\mathbf{t}}$, is
\end{center}
\begin{equation}
    \mathbf{E} = \mathbf{\mathbf{(e^{\xi_{\mathbf{R}}}R)^\top}[e^{\xi_{\mathbf{t}}}t]_{\times}}
\end{equation}
\begin{center}
The derivatives of $\xi_{\mathbf{R}}$ and $\xi_{\mathbf{t}}$ are taken at the identity, so they are equal to $0$ when treated as a constant. To compute the partial of this essential matrix w.r.t. $\xi_{\mathbf{R}}$ at $0$:
\end{center}
\begin{minipage}{.3\linewidth}
\begin{equation}
    \textbf{tx} \coloneq [\mathbf{t}]_\times \in \mathbb{R}^{3 \times 3}
\end{equation}
\end{minipage}
\begin{minipage}{.7\linewidth}
\begin{equation}
    \frac{\partial \mathbf{E}}{\partial \xi_{\mathbf{R}}} = \frac{\partial (e^{\xi_{\mathbf{R}}}\mathbf{R})^\top\mathbf{[t]_{\times}}}{\xi_{\mathbf{R}}} = \begin{bmatrix}
    \mathbf{R}^\top[\textbf{tx}_{c1}]_{\times} \\ 
    \mathbf{R}^\top[\textbf{tx}_{c2}]_{\times} \\ 
    \mathbf{R}^\top[\textbf{tx}_{c3}]_{\times}\\
    \end{bmatrix}^\top \in \mathbb{R}^{(3 \times 3) \times 3}
\end{equation}
\end{minipage}
\begin{center}
Similarly, the partial w.r.t. $\xi_{\mathbf{t}}$ at $0$:
\end{center}
\begin{equation}
    \frac{\partial \mathbf{E}}{\partial \xi_{\mathbf{t}}} = \frac{\partial \mathbf{R}^\top\mathbf{[e^{\xi_{\mathbf{t}}}t]_{\times}}}{\partial \xi_{\mathbf{t}}} = \frac{\partial \mathbf{R}^\top\mathbf{[e^{\xi_{\mathbf{t}}}t]_{\times}}}{\partial (e^{\xi_{\mathbf{t}}}\mathbf{t})} ~\frac{\partial (e^{\xi_{\mathbf{t}}}\mathbf{t})}{\partial \xi_{\mathbf{t}}} = \mathbf{R}^\top ~ \frac{\partial [\vec{n}]_{\times}}{\partial \vec{n}} ~ (-\textbf{tx}) \in \mathbb{R}^{(3 \times 3) \times 3}
\end{equation}
\begin{center}
    Putting it together with the chain rule:
\end{center}
\begin{align}
\label{eq:final_partials1}
\frac{\partial err(m,l)}{\partial \xi_{\mathbf{R}}} &= \frac{\partial err(m,l)}{\partial l} \frac{\partial l}{\partial \mathbf{F}} ~ \frac{\partial \mathbf{F}}{\partial \mathbf{E}} ~ \frac{\partial \mathbf{E}(e^{\xi_\mathbf{R}} \mathbf{R}, e^{\xi_{\mathbf{t}}} \mathbf{t})}{\partial \xi_{\mathbf{R}}} \in \mathbb{R}^{2 \times 3}\\
\label{eq:final_partials2}
\frac{\partial err(m,l)}{\partial \xi_{\mathbf{t}}} &= \frac{\partial err(m,l)}{\partial l} \frac{\partial l}{\partial \mathbf{F}} ~ \frac{\partial \mathbf{F}}{\partial \mathbf{E}} ~ \frac{\partial \mathbf{E}(e^{\xi_\mathbf{R}} \mathbf{R}, e^{\xi_{\mathbf{t}}} \mathbf{t})}{\partial \xi_{\mathbf{t}}} \in \mathbb{R}^{2 \times 3}\\
\end{align}

\section{Extracting Rotation and Translation from the Essential Matrix}

To obtain the relative pose given the essential matrix $\mathbf{E}$, we follow the procedure prescribed in ~\cite{cvbook}.\vspace{5mm}

\begin{minipage}{.3\linewidth}
\begin{equation}
W := \begin{bmatrix}
        0 & -1 & 0 \\
        1 & 0 & 0 \\
        0 & 0 & 1 \\
        \end{bmatrix}
\end{equation}
\end{minipage}
\begin{minipage}{.3\linewidth}
\begin{equation}
Z := \begin{bmatrix}
        0 & 1 & 0 \\
        -1 & 0 & 0 \\
        0 & 0 & 0 \\
        \end{bmatrix}
\end{equation}
\end{minipage}
\begin{minipage}{.3\linewidth}
\begin{equation}
    U, \Sigma, V^\top = \text{SVD}(\mathbf{E})
\end{equation}
\end{minipage}
\\\\\\\vspace{5mm}
\begin{minipage}{.3\linewidth}
\begin{equation}
t = UZU^\top
\end{equation}
\end{minipage}
\begin{minipage}{.3\linewidth}
\begin{equation}
R1 = UWV^\top
\end{equation}
\end{minipage}
\begin{minipage}{.3\linewidth}
\begin{equation}
R2 = UW^{\top}V^\top
\end{equation}
\end{minipage}
\\
The four plausible solutions are
\begin{equation}
    [(t, R1), (t, R2), (-t, R1), (-t, R2)]
\end{equation}
During training, we choose the solution closest to the ground-truth. During inference, we choose the pose which triangulates the most points in front of the camera.
\section{Architecture Details}

\begin{figure*}[h]
    \centering
    \includegraphics[width=0.7\textwidth]{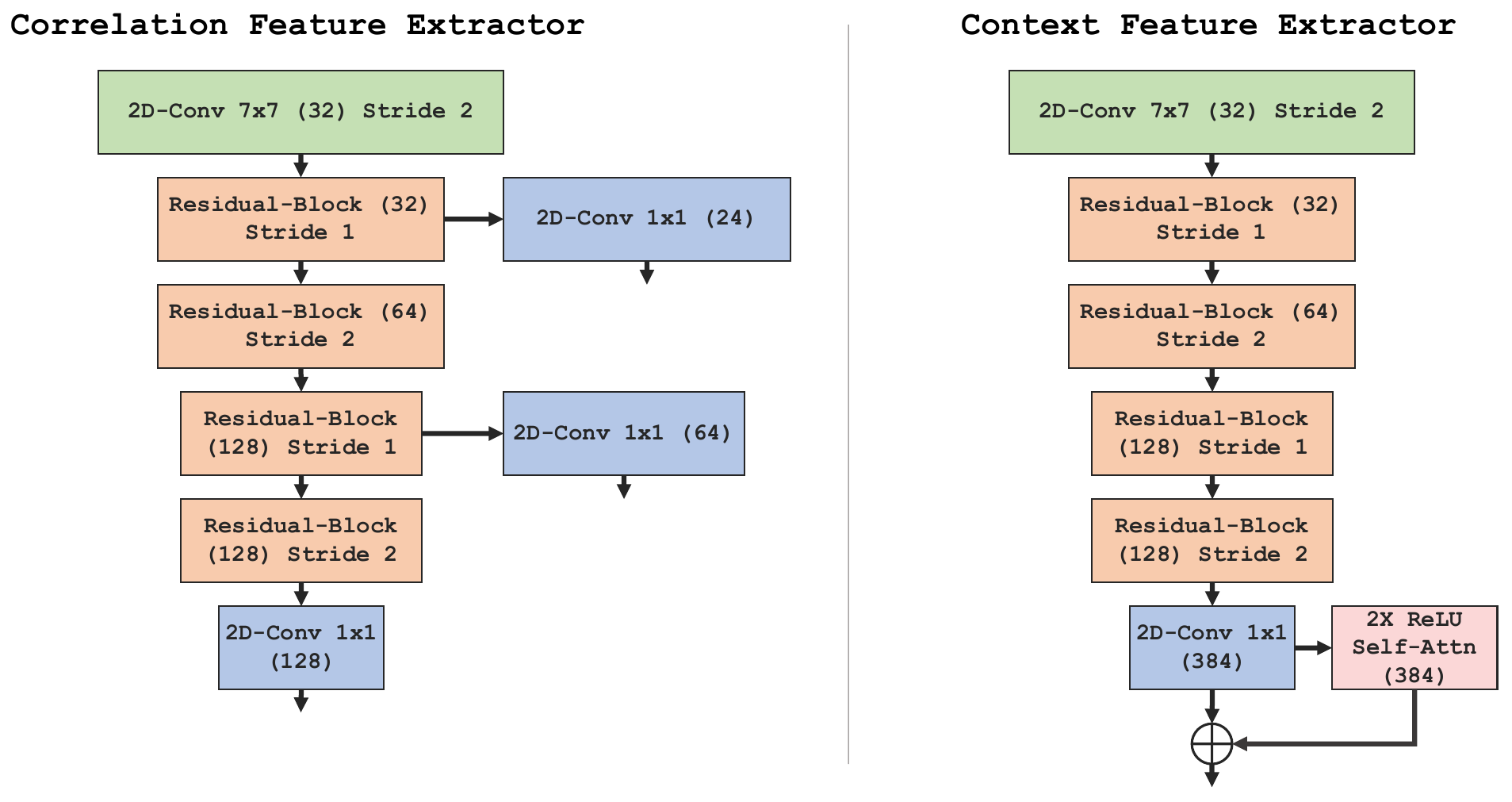}
    \caption{The architecture of our context and correlation feature extractors. Both feature extractors are residual networks. The context feature extractor also uses ReLU self-attention to propagate information across the image. The correlation features are used to evaluate visual similarity at multiple spatial resolutions. The numbers in parenthesis are the output feature dimensions.}
    \label{fig:feat_ext_arch}
\end{figure*}

\begin{figure*}[h]
    \centering
    \includegraphics[width=0.3\textwidth]{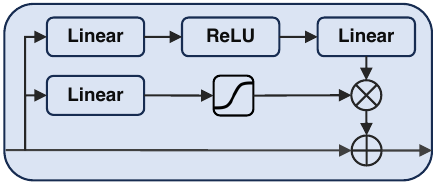}
    \caption{The gated residual unit.}
    \label{fig:mplgru}
\end{figure*}

\noindent\textbf{MLP-GRU} We depict the gated residual unit in Fig.~\ref{fig:mplgru}. This is the same design from DPVO~\cite{dpvo}.
\smallskip\\
\textbf{Feature Extractors} We visualize the feature extractors in Fig.~\ref{fig:feat_ext_arch}. The correlation features are produced at $\nicefrac{1}{2}$, $\nicefrac{1}{4}$ and $\nicefrac{1}{8}$ the image resolution. The context features are extracted only at $\nicefrac{1}{8}$ resolution, and have additional self-attention layers to propagate information over the image.%

\newpage

\section{Additional Qualitative Results}

\begin{figure}[h]
    \centering
    \includegraphics[width=\textwidth]{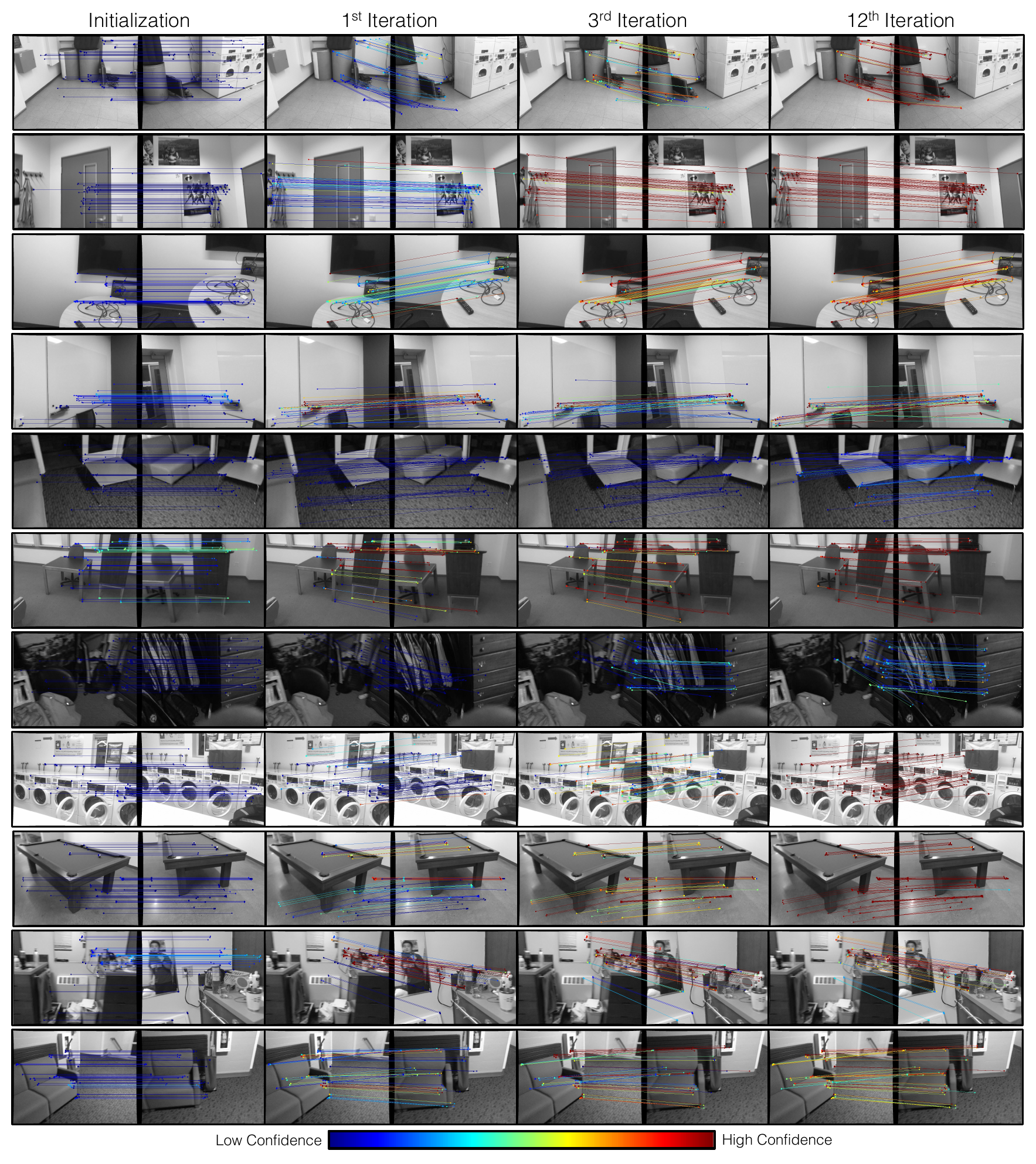}
    \caption{Additional Qualitative results on Scannet.}
    \label{fig:all_qual_iters}
\end{figure}

\end{document}